\documentclass[journal,twoside,web]{ieeecolor}
\usepackage{tmi}
\usepackage{cite}
\usepackage{amsmath,amssymb,amsfonts}
\usepackage{algorithmic}
\usepackage{graphicx}
\usepackage{textcomp}
\usepackage{amsmath}
\usepackage{adjustbox}
\usepackage{multirow}
\usepackage{array}
\usepackage{url}

\def\BibTeX{{\rm B\kern-.05em{\sc i\kern-.025em b}\kern-.08em
    T\kern-.1667em\lower.7ex\hbox{E}\kern-.125emX}}
\begin{document}
\title{Label-free Concept Based Multiple Instance Learning for Gigapixel Histopathology}
\author{Susu Sun, Leslie Tessier, Frédérique Meeuwsen, Clément Grisi, Dominique van Midden,\\ Geert Litjens, and Christian F. Baumgartner
\thanks{\textbf{This work has been submitted to the IEEE for possible publication. Copyright may be transferred without notice, after which this version may no longer be accessible.}}
\thanks{Funded by EXC number 2064/1 – Project number 390727645.
Corresponding author: Susu Sun (email: susu.sun@uni-tuebingen.de)}
\thanks{Susu Sun and Christian F. Baumgartner are with Cluster of Excellence: Machine Learning - New Perspectives for Science, University of Tübingen, Tübingen, Germany. Christian F. Baumgartner is also with the Faculty of Health Sciences and Medicine, University of Lucerne, Lucerne, Switzerland.}
\thanks{Leslie Tessier, Frédérique Meeuwsen, Clément Grisi, Dominique van Midden and Geert Litjens are with Radboud University Medical Center, Radboud, Netherlands. Leslie Tessier is also with Institut du Cancer de l’Ouest, Angers, France. Clément Grisi and Geert Litjens are also affiliated with the Oncode Institute, Utrecht, Netherlands}
}

\maketitle
\begin{abstract}
Multiple Instance Learning (MIL) methods allow for gigapixel Whole-Slide Image (WSI) analysis with only slide-level annotations. Interpretability is crucial for safely deploying such algorithms in high-stakes medical domains. Traditional MIL methods offer explanations by highlighting salient regions. However, such spatial heatmaps provide limited insights for end users. To address this, we propose a novel inherently interpretable WSI-classification approach that uses human-understandable pathology concepts to generate explanations. Our proposed Concept MIL model leverages recent advances in vision-language models to directly predict pathology concepts based on image features. The model's predictions are obtained through a linear combination of the concepts identified on the top-K patches of a WSI, enabling inherent explanations by tracing each concept's influence on the prediction. In contrast to traditional concept-based interpretable models, our approach eliminates the need for costly human annotations by leveraging the vision-language model. We validate our method on two widely used pathology datasets: Camelyon16 and PANDA. On both datasets, Concept MIL achieves AUC and accuracy scores over 0.9, putting it on par with state-of-the-art models. We further find that 87.1\% (Camelyon16) and 85.3\% (PANDA) of the top 20 patches fall within the tumor region. A user study shows that the concepts identified by our model align with the concepts used by pathologists, making it a promising strategy for human-interpretable WSI classification.
\end{abstract}


\section{Introduction}\label{sec:introduction}

\begin{figure*}[!t]
\centerline{\includegraphics[width=1.5\columnwidth]{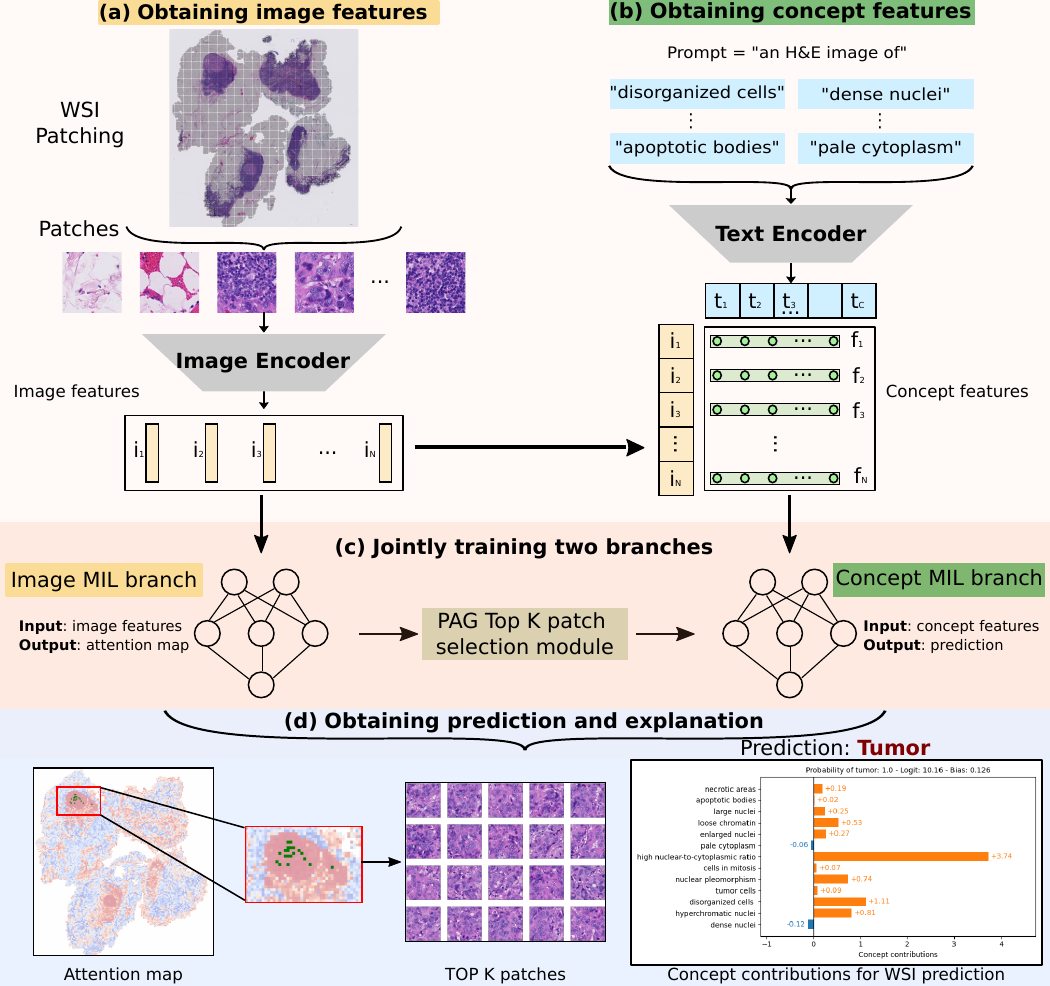}}
\caption{Training and obtaining prediction and explanation with Concept MIL. During training, we extract WSI image features as shown in \textbf{(a)} and project these image features to the predefined concept space to generate concept features as shown in \textbf{(b)}, then jointly train the image MIL branch and concept MIL branch through a patch selection module as shown in \textbf{(c)}. During inference, the concept MIL branch generates the final prediction using the concept features of top K patches selected by the image MIL branch. To explain the prediction for an individual sample, we provide the attention map from the image MIL branch with the top K patches highlighted by green dots, along with the corresponding top K patches and the concept contributions to the WSI prediction, as illustrated in \textbf{(d)}. 
}

\label{fig:framework}
\end{figure*}

\IEEEPARstart{A}{dvancements} in artificial intelligence, along with the growth of pathology datasets, have driven significant progress in histopathology \cite{srinidhi2021deep, campanella2019clinical, bejnordi2017diagnostic, Bulten2022}. 
However, there are several unique challenges in applying deep learning techniques directly to gigapixel whole-slide image (WSI) analysis. For instance, WSIs are high-resolution and thus large, making the end-to-end training of deep neural networks challenging\cite{dooper2023gigapixel, javed2022additive}. Moreover, pixel-level annotations are often unavailable, with only slide-level or patient-level weak annotations provided \cite{srinidhi2021deep, shao2021transmil}. To overcome these challenges, algorithms based on Multiple Instance Learning (MIL), such as Attention-based MIL\cite{ilse2018attention}, CLAM \cite{lu2021data}, and TransMIL\cite{shao2021transmil} have been developed, and have been used with great success in computational pathology. This type of model treats the patches from WSI as a bag of instances and trains models using only slide-level annotations in a weakly-supervised manner.

While these MIL approaches offer effective strategies for WSI analysis, there are concerns regarding their interpretability, which is essential for algorithms in safety-critical applications in medicine \cite{rudin2019stop}. MIL-based models typically learn an attention score for each patch to indicate its importance and provide an attention map as a spatial explanation for the prediction \cite{yao2020whole, lu2021data, li2021multi, javed2022additive}. Although attention maps have been shown to correlate with disease-relevant regions, there are several issues with using them as explanations\cite{javed2022additive}. The non-linear activations within MIL networks introduce a non-linear relationship between the attention scores and the final prediction, making this kind of explanation inaccurate and incapable of faithfully representing the model's real decision-making process\cite{javed2022additive, kapse2024si}.

Inherently interpretable models, such as those proposed in \cite{chen2019looks, koh2020concept, bohle2022b}, have recently gained attention in natural image analysis for their ``white box" property. This type of model enhances computational transparency and can provide explanations that faithfully reflect the model's underlying mechanisms. In the broader medical domain, several such models have been proposed for chest X-ray images\cite{sun2023inherently}, fundus images \cite{donteu2024sparse}, and biological sequences\cite{ditz2023inherently}. However, there is relatively little work in computational pathology, particularly for WSI analysis. To our knowledge, only two inherently interpretable models for WSI analysis exist: Additive MIL\cite{javed2022additive} and Self-Interpretable MIL (SI-MIL) \cite{kapse2024si}. Notably, SI-MIL \cite{kapse2024si} achieves inherent interpretability by providing explanations using pathology-related features, such as the statistical properties of nuclei. While these explanations deliver more information than attention maps, the pathology-related features are hand-crafted and derived from the outputs of a nuclei segmentation model pre-trained on other datasets. This approach limits the model's flexibility to adapt to different diseases and can inherit errors from the segmentation model. Furthermore, explanations from SI-MIL are based on 205-dimensional pathology-related features, which remain challenging for humans to interpret.

Concept-based interpretable models such as Concept Bottleneck Models (CBMs) \cite{koh2020concept, oikarinen2023label, yuksekgonul2022post} provide explanations using human-understandable concepts, making them an increasingly important category in explainable AI. However, these models typically rely on concept annotations, restricting the concepts to those already known by the annotator and requiring expert input. Recently, researchers started to build CBMs in a label-free way \cite{yuksekgonul2022post, oikarinen2023label} by leveraging vision-language models such as CLIP \cite{radford2021learning}, which are pretrained on large-scale image-caption datasets. This approach removes the need for manual labeling and obtains promising results on the natural images. In the field of computational pathology, vision-language foundation models such as PLIP\cite{huang2023visual} and CONCH \cite{lu2024avisionlanguage}, which are specifically trained on histopathology data, have been proposed recently and have significantly enhanced the performance of downstream tasks. These foundation models offer a tool for building concept-based models in a label-free manner. However, to our knowledge, there is no concept-based interpretable model for WSI analysis.

Here, we introduce Concept MIL, a concept-based, inherently interpretable MIL model for WSI classification. Our model is influenced by SI-MIL \cite{kapse2024si} and incorporates recent advancements in concept-based interpretable models. We address the limitations of SI-MIL by replacing the self-interpretable MIL branch with one that employs flexible, pathologist-defined concepts. This approach removes the potential mistakes from upstream models, reduces the tedious task of designing and extracting pathological features, and offers greater flexibility for deployment across various diseases. Our Concept MIL delivers faithful local and global explanations through pathological concepts, ensuring high interpretability while maintaining robust performance. Importantly, our model eliminates the need for manual concept labeling by leveraging the pathology vision-language foundation model CONCH \cite{lu2024avisionlanguage}, offering flexibility in model design and easy adaptions across various diseases. We conduct both quantitative and qualitative evaluations of the model’s local and global explanations, including a user study with pathologists. The results demonstrate that Concept MIL achieves classification performance comparable to state-of-the-art MIL models, accurately identifies disease-relevant regions, and provides explanations aligned with clinical knowledge.
The code, data and user study questionnaire will be made publicly available at \underline
{https://github.com/ss-sun/ConceptMIL} upon acceptance.

\section{Related Works}\label{sec:releated_work}

\subsection{Interpretable Methods for Histopathology}\label{sec:inter_methods}

Interpretability is crucial for machine learning algorithms in the medical domain. Post-hoc explanation methods generate insights by applying approximations to trained models. In histopathology, post-hoc explanation techniques such as LIME, Shapley values, and gradient-based methods were used to provide explanations for a model's decisions on tumor grading\cite{wang2019machine}, supporting the validations of disease biomarkers discovery \cite{huang2023explainable}, and explaining WSI classification \cite{pirovano2020improving}. These methods are model-agnostic, making them generalize well to different models. However, the explanations they generate have been shown to lack dependence on both the training data and the trained model, leading to unfaithful representations of the model's true decision mechanism \cite{adebayo2018sanity, javed2022additive}. 

Multiple instance learning (MIL) models in computational pathology often directly provide interpretability by leveraging internal computations, such as learned attention scores, to identify regions that contribute to predictions~\cite{yao2020whole, lu2021data, li2021multi, javed2022additive}. While these explanations can be directly obtained during inference without applying post-hoc methods, they still suffer from several limitations. Due to the non-linear activations within MIL networks, patches with low attention scores can still contribute substantially to the final prediction. Therefore, attention maps fail to faithfully represent the model’s true decision-making mechanism, undermining their reliability as explanations, as thoroughly discussed in \cite{javed2022additive}. 

Inherently interpretable models have recently gained attention for their capability to provide faithful explanations\cite{rudin2019stop}. In the medical field, these models have been explored to offer more reliable and trustworthy explanations \cite{wei2024mprotonet, sun2023inherently, donteu2024sparse}. However, relatively little work has been done in the area of WSI analysis in histopathology. To our knowledge, Additive MIL\cite{javed2022additive} and Self-Interpretable MIL (SI-MIL) \cite{kapse2024si} are the only two inherently interpretable models for WSI classification. Additive MIL \cite{javed2022additive} achieves inherent interpretability by introducing a novel formulation of MIL that allows an exact decomposition of the model's predictions in terms of spatial regions of inputs. Since the explanations are presented in a traditional heatmap style that highlights important regions, Additive MIL offers limited insight for end users. SI-MIL focuses on user-friendly explanations by presenting to the user how pathological features, such as the nuclei morphology and the spatial distribution of cells, influence the prediction. Although SI-MIL is performant and inherently interpretable, the pathological features it uses are hand-crafted statistical properties of nuclei, derived based on the outputs of a HoVerNet \cite{graham2019hover} model pre-trained on the PanNuke\cite{gamper2019pannuke} dataset. Consequently, SI-MIL inherits misclassifications from HoVerNet. Additionally, the use of 205 morphometric properties as pathological features adds complexity to feature extraction and challenges human interpretability.

\subsection{Concept-based Interpretable Models}\label{sec:concept_based_related_work}

Current explanation methods in histopathology primarily rely on low-level features, such as image features of pixels or patches, to provide importance scores that highlight key features. However, systematically summarizing and interpreting these scores is challenging and varies among users, which can easily introduce human confirmation biases \cite{kim2018interpretability, ghorbani2019towards}. 
Another line of research has focused on providing explanations using high-level concepts that are more human-interpretable \cite{kim2018interpretability, ghorbani2019towards}, for example, using concepts such as ``strip" and ``long tail" to detect a zebra. These concept-based explanations are directly built on human knowledge, making them intuitive and easy to understand. Such approaches have been explored in the histopathology domain. Graziani et al.\cite{graziani2018regression} introduced the Regression Concept Vector (RCVs) to explain deep neural network predictions using domain-specific concepts such as nuclei size. Pinckaers et al. \cite{pinckaers2022predicting} applied concept-based explanation methods to interpret the tissue patterns learned by a deep neural network. However, these methods rely on post-hoc mechanisms, which can lead to unfaithful representations of the model’s true decision-making process.

Concept Bottleneck Models (CBMs) \cite{koh2020concept, oikarinen2023label, yuksekgonul2022post}, on the other hand, are inherently interpretable models that offer explanations using concepts. This type of model achieves inherent interpretability by first detecting concepts from the input and then using linear combinations of these highly interpretable concepts to make final predictions and provide explanations. CBMs have been applied in the medical domain. For instance, Fauw et al. \cite{de2018clinically} employed CBM for diagnosis and referral in retinal disease, and Koh et al. \cite{koh2020concept} used them for grading knee x-rays. 
While CBMs can be trained end-to-end with supervision on both concepts and classes, their practical application is often limited by the need for human annotations of concepts, which can be costly and time-consuming. Newer CBM models such as Label-free CBM \cite{oikarinen2023label}, DN-CBM \cite{rao2024discover}
allow mapping inputs to concepts without a training set of labeled concepts by leveraging the shared embedding space in the vision-language models such as CLIP\cite{radford2021learning}. In histopathology, vision-language foundation models like PLIP \cite{huang2023visual} and CONCH \cite{lu2024avisionlanguage} have been recently proposed and have shown promising results in various tasks. In this work, we use the vision-language model CONCH\cite{lu2024avisionlanguage} to develop a concept-based WSI classification model that does not require labeled concepts. To our knowledge, our proposed Concept MIL model is the first concept-based, inherently interpretable model for WSI analysis.

\section{Methods}\label{sec:method}

In the following, we discuss the main components of our approach. Our model consists of two branches: the image MIL branch, which identifies the top K most significant patches, and the concept MIL branch, which generates final predictions and explanations using the concept activation vectors of the selected patches (see Fig.~\ref{fig:framework}). First, we explain the image features extraction from the WSI (Fig.~\ref{fig:framework}a, Sec.~\ref{sec:method_image_features}). Next, we describe how to project the image features to a set of predefined pathology concepts and obtain the concept activation vectors (Fig.~\ref{fig:framework}b, Sec.~\ref{sec:method_concepts_features}). Then, we explain the joint training of our dual-branch Concept MIL model (Fig.~\ref{fig:framework}c, Sec.~\ref{sec:method_training}). Finally, we describe how the model generates predictions and inherent explanations (Fig.~\ref{fig:framework}d, Sec.~\ref{sec:method_local_x}, Sec.~\ref{sec:method_global_x}).

\subsection{Generating Image Features}\label{sec:method_image_features}

In the image feature generation step, we extract patch-wised image features from the WSI. As shown in Fig.~\ref{fig:framework}a, the tissue regions are segmented from the WSI and then cropped into $N$ non-overlapping patches. Afterward, we employ the image encoder from the histopathology-specific vision-language model CONCH \cite{lu2024avisionlanguage} to extract image features for these patches. For simplicity, we extract patches of the same size $256 \times 256$ at the resolution level of  $0.5\mu m / \text{pixel}$. 

After the image feature extraction step, the $N$ patches from WSI are converted into $N$ image embedding vectors $i_1, i_2 ... i_N$, each with a dimensionality of $D=512$. These image features are subsequently used for generating concept-based features, as depicted in Fig.~\ref{fig:framework}b, and for training a conventional image MIL branch, as shown in Fig.~\ref{fig:framework}c.

\subsection{Generating Concepts Features}\label{sec:method_concepts_features}

As illustrated in Fig.~\ref{fig:framework}b, the concept-based features $f_n$ are computed based on the image features $i_n$ from Fig.~\ref{fig:framework}a and text embeddings $t_c$ extracted from the target concept set. We use the text encoder from the vision-language model CONCH\cite{lu2024avisionlanguage} to generate text embeddings. Since CONCH is trained on large-scale histopathology image-text pairs, it learns a shared embedding space for images and text, enabling the direct prediction of text-based concepts from image features. This approach eliminates the need for concept annotations in the concept feature generation step. 

Following the way of Label-free CBMs \cite{oikarinen2023label}, we initialize a concept set by prompting GPT4o \cite{openai2023chatgpt} for features of tumor tissue. For instance, to collect pathology concepts for detecting breast cancer tumor tissues, we ask GPT4o the following question: ``List the most important features for recognizing breast cancer metastases in hematoxylin and eosin (H \& E) stained lymph node whole-slide images". We additionally ask a pathologist to refine the initialized concept set to ensure the quality. In routine practice, pathologists examine whole slide images for morphological features across different resolutions and scales to make diagnoses. However, our model currently focuses on features at a specific scale and resolution. To make sure that the defined concepts are actually visible at this scale, we provide both GPT4o and pathologists with a few image patches as references during the initialization and adjustment of the concepts. This allows us to obtain a set of concepts that are relevant and visible at the given resolution and scale.

To generate the text embeddings for the target concepts, we construct the prompt as ``an H \& E image of CONCEPT'' by replacing the CONCEPT placeholder with a specific concept in the target concept set. These prompts are projected into text embeddings using the CONCH\cite{lu2024avisionlanguage} text encoder. The image features $i_n$ are then mapped to the concept space by calculating the cosine similarity between the normalized image embeddings and text embeddings, producing the concept activation vectors $f_n$ for the $N$ patches.

\begin{figure}[!t]
\centerline{\includegraphics[width=0.8\columnwidth]{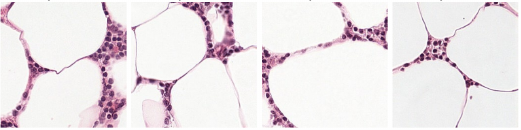}}
\caption{Patches containing fat cells get high cosine similarity scores with the concept of ``fibrous tissue'', indicating a potential misalignment between the image and text spaces in CONCH\cite{lu2024avisionlanguage}.}
\label{fig:top_patches}
\end{figure}

In our preliminary experiment, we observed misalignments between the image and text spaces in the CONCH model. For instance, as shown in Fig.~\ref{fig:top_patches}, patches containing fat cells get high cosine similarity scores with the concept ``fibrous tissue'', indicating that CONCH may not fully understand the concept ``fibrous tissue'' in the context of histopathology. Such misalignments can affect user's trust in the model and explanations. To address this, we conduct an additional validation step to filter out misaligned concepts. We select 50K patches from tumor regions based on ground truth masks from the training set. For each concept, we retrieve five patches with high cosine similarity scores and ask a pathologist to verify the relevance of the target concept in the selected patches. Since the target concept sets are relatively small, this verification process is easy to perform. Finally, we use the refined concepts listed in Table \ref{tab:conepts} for concept activation vector extraction.

\subsection{Training}\label{sec:method_training}

Similar to the SI-MIL\cite{kapse2024si} model, we train the image MIL branch and the self-interpretable concept MIL branch jointly through a Patch Attention-Guided (PAG) Top-K module (see Fig.~\ref{fig:framework}c). 

We implement the image MIL branch in Fig.~\ref{fig:framework}c as a conventional attention-based MIL model. It treats $N$ image features from a WSI as a bag of instances, transforms them with a projector $H(\cdot)$, and weights the projected features with patch-wise attention scores $\alpha$ from attention module $A^p (\cdot)$. The transformed image features and the attention scores are denoted as:

\begin{equation}
    V = H(I); \quad \alpha = A^p(V).
\end{equation}

Here, $I$ represents the input image feature matrix for the WSI, where each individual image feature is $i_n$. $V$ represents the transformed image features, with each feature vector represented as $v_n$. $\alpha$ refers to the attention scores, with $\alpha_n$ being the attention score for patch $n$.

The prediction, denoted as $\hat{Y}_{img}$, is obtained by applying logistic regression to the scaled image features, as follows:

\begin{equation}
    \hat{Y}_{img} = \sigma \left(\sum_{n=1}^{N} \sum_{d=1}^{D} w_d^{\prime} \cdot(v_{nd}\cdot \alpha_n) + b_{img}\right),
\end{equation}
where $\sigma$ represents the logistic function, $v_{nd}$ denotes the value of the $d$-th dimension of the transformed image feature $v_n$, and $w_d^{\prime}$ and  $b_{img}$ 
correspond to the weights and bias of the linear classifier, respectively.

The PAG Top-K module identifies the top K patches based on the highest attention scores in $\alpha$ and passes the indices of the selected patches to the concept MIL branch. Instead of using non-differentiable standard Top-K selection, we adopt the differentiable PAG Top-K module used by \cite{kapse2024si, thandiackal2022differentiable}. This module implements a differentiable Top-K operator based on the perturbed maximum method \cite{cordonnier2021differentiable, thandiackal2022differentiable}. During the forward pass, uniform Gaussian noise is added to each attention score to create perturbed attention values, and the corresponding linear program for maximization is solved. In the backward pass, the Jacobian for the forward operation is computed. This differentiable Top-K selection allows learning the parameters of the attention module in the image MIL branch, thus enabling the joint training of two branches.

The concept MIL branch in Fig.~\ref{fig:framework}c receives the concept activation vectors $f_j$ of the top K salient patches selected by the PAG Top-K module and makes predictions based on these concept-based features. The concept attention scores $\beta$ are first calculated using the attention module $A^c(\cdot)$, scaled by its $\gamma ^{th}$ percentile $Pr_{\gamma}$, and then transformed using a sigmoid function with a temperature hyperparameter $t$. This operation turns the attention scores that are less than the $\gamma ^{th}$ percentile towards zeros, enforcing sparsity in feature selection. 
The resulting concept attention scores $\beta$ are given by:

\begin{equation}
    \widetilde{\beta} = A^c(F^T)
\end{equation}

\begin{equation}
    \beta_\text{scaled} = \frac{\widetilde{\beta} - Pr_{\gamma}}{std(\widetilde{\beta})}; \quad \beta= \frac{1}{1 + e^{-\beta_\text{scaled} \cdot t}},
\end{equation}
where $F^T$ is the transposed concept-based feature matrix of top K patches, $\widetilde{\beta}$, $\beta_\text{scaled}$, and $\beta$ denote the raw concept attention scores from the attention module, the scaled scores, and the transformed scores, respectively. $std (\cdot)$ is the standard deviation, and $t$ is a scaling hyperparameter.

The transformed concept attention scores $\beta$ are applied to scale the corresponding concept activation vectors, and the final prediction of the concept MIL branch is a linear combination of $C$ concepts from the top $K$ patches:

\begin{equation}
    \hat{Y}_{concept} = \sigma \left(\sum_{j=1}^{K} \sum_{c=1}^{C} w_c \cdot (f_{jc} \cdot \beta_c ) + b_{concept}\right), \label{eq:concept_pred}
\end{equation}
where $\sigma$ is a sigmoid function, $f_{jc}$ represents the value of the $c$-th concept in the concept activation vector $f_j$, $\beta_c$ represents the attention score for concept $c$, and $w_c$ and $b_{concept}$ are the weights and bias terms of a linear classifier.

During training, both the image MIL branch and the self-interpretable concept MIL branch perform classification, generating predictions $\hat{Y}_{img}$ and $\hat{Y}_{concept}$. We use the binary cross-entropy (BCE) loss to calculate the classification loss. In addition, we apply an $L_2$ regularization to further constrain the attention scores  $\alpha$ from the image MIL branch. We optimize this dual branch MIL model with the following loss function:

\begin{equation}
L = L_{BCE}(Y, \hat{Y}_{img}) + L_{BCE}(Y, \hat{Y}_{concept}) + \lambda L_2(\alpha),
\label{eq:overall_loss}\end{equation} 
where $Y$ is the ground truth of the WSI, and $\lambda$ is a hyper-parameter for the $L_2$ regularization term.

\subsection{Obtaining Prediction and Local Explanation}
\label{sec:method_local_x}

\begin{figure*}[!t]
% \centerline{\includegraphics[width=1.8\columnwidth]{LaTeX/figs/local_exp.png}}
\centerline{\includegraphics[width=1.5\columnwidth]{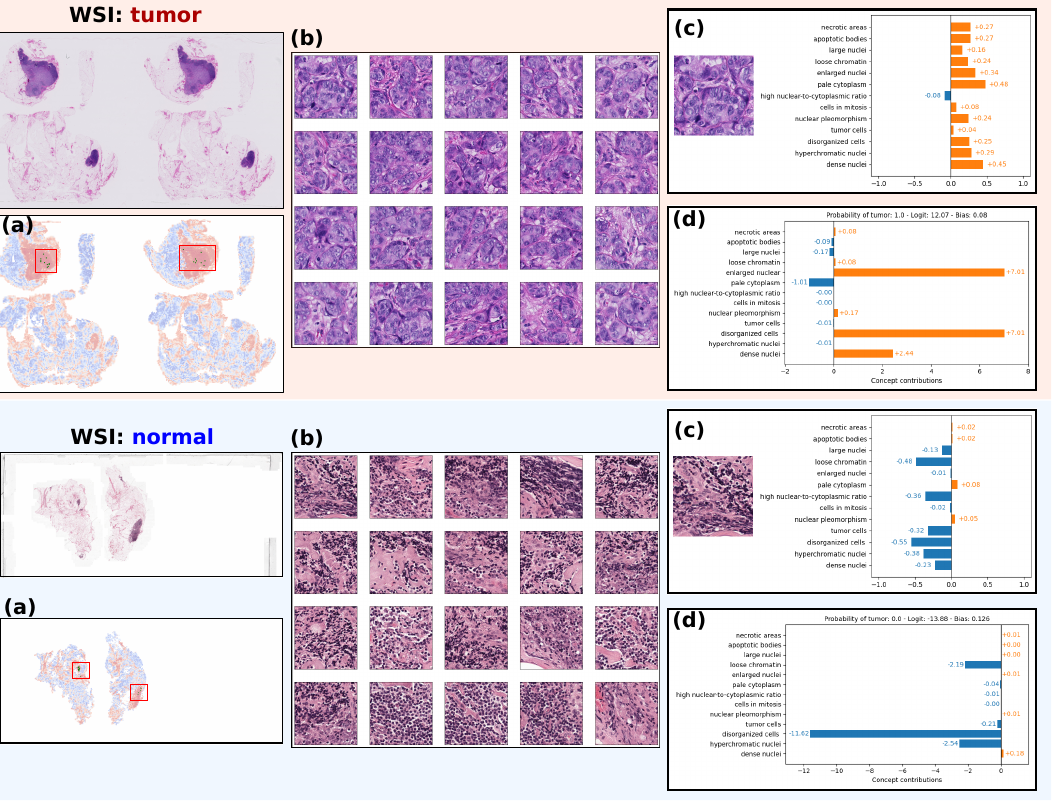}}
\caption{Local explanations for predictions of tumor and normal cases from Camelyon16 dataset. A local explanation for a WSI prediction includes four components: \textbf{(a)} an attention map, \textbf{(b)} the top 20 patches selected based on attention scores, \textbf{(c)} concept features for each selected patch, and \textbf{(d)} the whole slide level concept contribution vector.}
\label{fig:local_exp}
\end{figure*}

During inference, we discard the prediction from the image MIL branch and only use the prediction $\hat{Y}_{concept}$ from the concept MIL branch as the final output, ensuring the prediction is inherently interpretable. 

As shown in \eqref{eq:concept_pred}, $\hat{Y}_{concept}$ is a linear combination of concept activation vectors from the top K patches,
the contribution of a specific concept to the WSI prediction can be written as:

\begin{equation}
\kappa_c = w_c \cdot \sum_{j=1}^{K}f_{jc}  \cdot  \beta_c . 
\end{equation}

Thus, the prediction $\hat{Y}_{concept}$ can be reformulated as the sum of the contributions from each concept, along with a bias:

\begin{equation}
\hat{Y}_{concept} = \sigma \left(\sum_{c=1}^{C} \kappa_c + b_{concept}\right).
\end{equation}

For a WSI prediction $\hat{Y}_{concept}$, we provide a local explanation consisting of four components, as shown in Fig.~\ref{fig:local_exp}. The first component is the attention map $\alpha$ (Fig.~\ref{fig:local_exp}a) from the image MIL branch, highlighting the top K patches that indicate the key regions influencing the prediction. The second component is a visualization of the top K patches (Fig.~\ref{fig:local_exp}b), allowing the end user to examine the patches and inspect the image details. The third component includes the concept activation vectors for each patch (Fig.~\ref{fig:local_exp}c), enabling a detailed exploration of individual concept-based features. Finally, the fourth component is the WSI-level concept contributions (Fig.~\ref{fig:local_exp}d), where the contribution $\kappa_c$ of each concept to the final prediction can be assessed.

\subsection{Obtaining Global Explanations}\label{sec:method_global_x}

\begin{figure}[!t]
\centerline{\includegraphics[width=0.95\columnwidth]{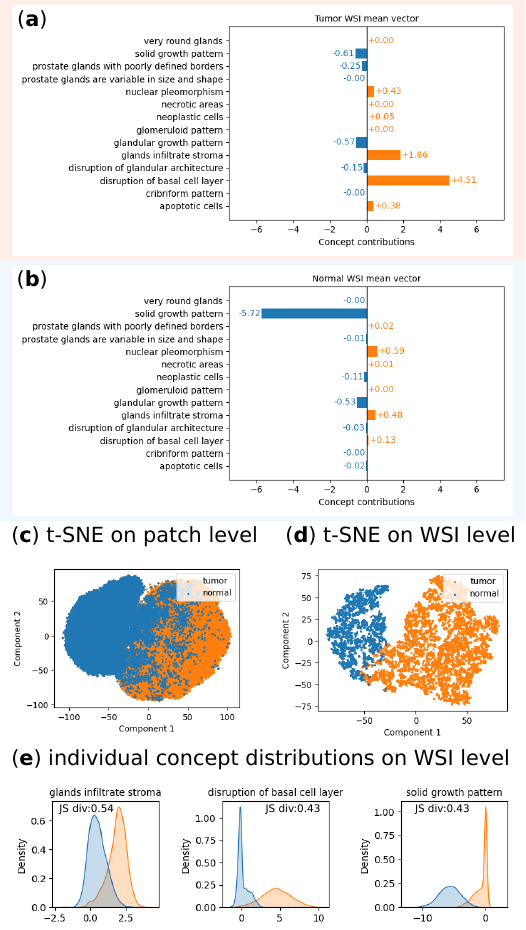}}
\caption{Global explanations for the model trained on PANDA dataset.
\textbf{(a)} and \textbf{(b)} are WSI-level mean concept contribution vectors for tumor and normal predictions. \textbf{(c)} and \textbf{(d)} show t-SNE plots of concept vectors from normal and tumor cases at the patch level and WSI level. \textbf{(e)} shows distributions of three individual concepts across normal and tumor cases, along with their corresponding Jensen–Shannon divergence scores.}
\label{fig:global_exp}
\end{figure}

In addition to local explanations for individual samples, our model offers a global explanation of its overall prediction mechanism. Figure~\ref{fig:global_exp}a and b show the mean WSI-level concept contributions for tumor and normal samples in the entire train set. These mean concept contribution vectors provide a dataset-level perspective on how each concept, on average, influences the tumor or normal prediction. Beyond the WSI-level mean vectors, we can project the concept vectors into 2D space with t-SNE plot \cite{van2008visualizing} to explore how well in the feature space the tumor and normal cases are separated, as shown in Fig.~\ref{fig:global_exp}c and d.
Additionally, we can look into the distributions of a specific concept across tumor and normal cases at both the patch and WSI levels, as illustrated in Fig.~\ref{fig:global_exp}e. These global explanations provide a systematic approach to understanding the model, allowing for an evaluation of the model's quality and identifying potential directions for further improvement.

\section{Experiments and Results}

\subsection{Datasets}\label{datasets}

We present results on two widely-used computational pathology datasets: Camelyon16 \cite{bejnordi2017diagnostic} and PANDA \cite{Bulten2022}. Camelyon16 is a breast cancer metastasis detection dataset consisting of 270 WSIs in the training set and 129 samples in the test set labeled as tumor or normal. We split the official training set into training and validation subsets with ratios of 0.8 and 0.2, while the official test set was used for evaluation. PANDA is a prostate cancer grading dataset comprising 10,616 digitized prostate biopsies in the official training set. Each WSI is annotated with Gleason scores\cite{epstein2010update} and International Society of Urological Pathology (ISUP) grades to reflect cancer severity, which can be further mapped to binary class labels of tumor and normal. Since the official PANDA test set is not publicly available, we divided the training set into training, validation, and test subsets with ratios of 0.8, 0.1, and 0.1, respectively. Further, ground-truth tumor masks are available for all WSIs in Camelyon16 and most in PANDA. We used those pixel-level labels to quantitatively evaluate the attention map from the local explanations.

\subsection{Implementation details}\label{implementation}

For all WSIs, we generated image and concept features and trained our Concept MIL model as described in Sec.~\ref{sec:method}. We implemented the image MIL branch in our model as attention-based MIL \cite{ilse2018attention}. The image feature projector $H(\cdot)$ was implemented as a Fully Connected Layer (FCL) followed by the activation function ReLU. The attention module $A^p (\cdot)$ was a gated attention module following \cite{lu2021data}. We used an FCL followed by a sigmoid function for the classification in the image branch. The PAG Top K patch selection module was adapted from \cite{kapse2024si, thandiackal2022differentiable}, and we set K to 20 for models on both datasets. In the concept MIL branch of our model, the attention module $A^c (\cdot)$ was also implemented as a gated attention mechanism, and the classification layer consisted of a fully connected layer with a sigmoid function. We set percentile $Pr_{\gamma}$ to 0.75 and temperature $t$ to 3 for scaling the attention scores $\beta$. And set the $\lambda$ in loss function \eqref{eq:overall_loss} to 0.05.

During training, we used a batch size of 1 to fit the WSIs with different bag sizes. The learning rate was set to 0.001 for the Camelyon16 dataset and 0.0001 for the PANDA dataset, with a weight decay of 0.001 applied to both datasets. The model was trained for 300 epochs, and the final model from the last epoch was used for evaluation.

\begin{table}
\caption{Concept sets for CAMELYON16 and PANDA}
\label{tab:conepts}
\centering
\begin{tabular}{>{\centering\arraybackslash}p{90pt}|>{\centering\arraybackslash}p{120pt}}
        \hline
        \textbf{Camelyon16} & \textbf{PANDA} \\
        \hline
\end{tabular}
\begin{tabular}{>{\raggedright\arraybackslash}p{90pt}|>{\raggedright\arraybackslash}p{120pt}}
\textbullet \hspace{2pt}dense nuclei & \textbullet \hspace{2pt}apoptotic cells\\
\textbullet \hspace{2pt}hyperchromatic nuclei & \textbullet \hspace{2pt}cribriform pattern \\
\textbullet \hspace{2pt}disorganized cells & \textbullet \hspace{2pt}disruption of basal cell layer \\
\textbullet \hspace{2pt}tumor cells & \textbullet \hspace{2pt}glandular growth pattern \\
\textbullet \hspace{2pt}nuclear pleomorphism & \textbullet \hspace{2pt}glands infiltrate stroma \\
\textbullet \hspace{2pt}pale cytoplasm & \textbullet \hspace{2pt}glomeruloid pattern \\
\textbullet \hspace{2pt}enlarged nuclei & \textbullet \hspace{2pt}neoplastic cells \\
\textbullet \hspace{2pt}loose chromatin & \textbullet \hspace{2pt}necrotic areas \\
\textbullet \hspace{2pt}large nuclei & \textbullet \hspace{2pt}nuclear pleomorphism \\
\textbullet \hspace{2pt}necrotic areas & \textbullet \hspace{2pt}solid growth pattern \\
\textbullet \hspace{2pt}apoptotic bodies & \textbullet \hspace{2pt}very round glands\\
\textbullet \hspace{2pt}cells in mitosis & \textbullet \hspace{2pt}disruption of glandular architecture \\
\textbullet \hspace{2pt}high nuclear-to- cytoplasmic ratio& \textbullet \hspace{2pt}prostate glands with poorly defined borders\\
& \textbullet \hspace{2pt} prostate glands are variable in size and shape\\
\hline
\end{tabular}
\end{table}

\subsection{Baselines}

We compared our model with Attention MIL \cite{ilse2018attention}, CLAM \cite{lu2021data}, TransMIL \cite{shao2021transmil} and Additive MIL \cite{javed2022additive}. We used the same image features extracted with CONCH\cite{lu2024avisionlanguage} as input for all models. The implementation of the baseline models was based on their official code releases. Unfortunately, a direct comparison to SI-MIL was not possible. As discussed in Sec.~\ref{sec:inter_methods}, the pathological features used by SI-MIL\cite{kapse2024si} rely on the output from HoVerNet \cite{graham2019hover} which cannot be directly applied to the datasets used in this paper. A comparison to the values reported in the SI-MIL publication is also not feasible, as no quantitative evaluations of the local explanations were conducted. Additionally, the reported Jensen–Shannon divergence values in \cite{kapse2024si} for the global explanations are in a different range and cannot be directly compared to ours.

We trained Attention MIL, TransMIL, and Additive MIL for 300 epochs with a learning rate of 0.0002 and weight decay of 0.001. We trained the CLAM multi-branch model for 200 epochs with a learning rate of 0.0001 following the official code release. For all compared models, we used the models from the last epoch for evaluation.

\subsection{Metrics}\label{sec:metrics}

\begin{figure}[!t]
\centerline{\includegraphics[width=0.75\columnwidth]{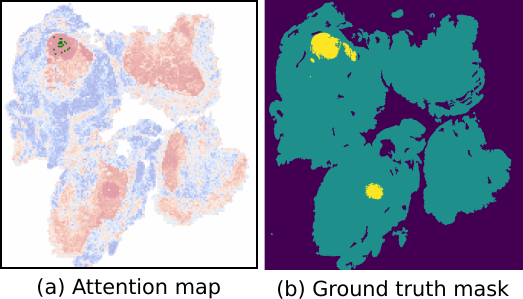}}
\caption{\textbf{(a)} Attention map from Concept MIL model on a Camelyon16 sample, with the top 20 patches highlighted by green dots. \textbf{(b)} Ground truth mask for the sample, showing tumor regions in yellow and normal tissue in dark green.}
\label{fig:metric_hits}
\end{figure}

We evaluated our model and the baseline models on the classification performance and explanation quality. We used Accuracy and the Area Under the ROC Curve (AUC) as the evaluation metrics for classification. To quantitatively assess local explanations,  we defined disease localization score based on the pointing game \cite{zhang2018top} to evaluate how well the salient patches locate the tumor regions. The idea of pointing game is to check whether the most salient point falls inside the ground truth object regions. We extended this approach by locating the K patches with the highest attention scores and checking if they are inside the ground truth tumor regions, as illustrated in Fig.~\ref{fig:metric_hits}. The disease localization score was defined as follows:

\begin{equation}
\text{Disease Localization Score} = \frac{\text{ Hits@K}}{K},
\end{equation}
where \text{Hits@K} represents the number of top K patches that fall within the ground truth tumor regions. We set K to 20 for all evaluated explanations.

We quantitatively evaluated global explanations in terms of the class separability of the learned features \cite{kapse2024si, fremond2023interpretable}. We used the Jensen–Shannon (JS) divergence to quantify the difference between the distributions of tumor and normal samples. Additionally, we projected the concept vectors to 2D space using t-SNE\cite{van2008visualizing} plot and calculated the Silhouette score \cite{rousseeuw1987silhouettes} to evaluate how well the tumor and normal samples were clustered. Since the baseline models do not provide global explanations, we performed this evaluation only on our model.

\subsection{Classification Performance}

The classification results of all compared models are presented in Table \ref{tab:eval_classification}. Our model achieved classification performance on par with state-of-the-art baselines, demonstrating that it maintains classification performance while offering interpretability. We conducted an ablation study on the two MIL branches by training them separately, using only image features or concept-based features. The ablation results showed that using only a concept MIL branch did not yield strong classification performance, indicating that guidance from the image MIL branch is necessary for effective classification.  

\begin{table}
    \caption{Classification performance measured by Accuracy and AUC. }
    \label{tab:eval_classification}
    \centering
    \begin{tabular}{c|cc|cc}
        \hline
         \textbf{Model} & \multicolumn{2}{c|}{\textbf{Camelyon16}} & \multicolumn{2}{c}{\textbf{PANDA}} \\
         & Acc. & AUC & Acc. & AUC \\
        \hline
        ABMIL &0.914 & 0.929 & 0.917 & 0.972 \\
        TransMIL & 0.938 & 0.950 & 0.934 & 0.969\\
        AdditiveMIL & 0.891 & 0.856 & 0.920 & 0.971\\
        CLAM & 0.906 & \textbf{0.978} & 0.924 & \textbf{0.982} \\
        ours (image)  & \textbf{0.938} & 0.970 & \textbf{0.937}& 0.979\\
        ours (concept) & 0.623& 0.688 & 0.900& 0.955\\
        ours & 0.898 & 0.916  & 0.927 & 0.977 \\
        \hline
    \end{tabular}
\end{table}

\subsection{Evaluations of Local Explanations}

Several MIL models offer local explanations through attention scores-based heatmaps. However, these explanations are often assessed qualitatively. In this work, we conduct both quantitative and qualitative evaluations of local explanations.

\subsubsection{Quantative Evaluation}

We evaluated the local explanations from our model and baselines using the disease localization score defined in Sec.~\ref{sec:metrics}. We excluded TransMIL\cite{shao2021transmil} from this evaluation because the official paper lacks details on generating explanations for it. The disease localization scores in Table \ref{tab:disease_localization} show that the top 20 patches selected by the MIL models effectively identified the tumor regions. Our model achieved the highest localization score on the Camelyon16 dataset and the second-best on the PANDA dataset, with an average of 17 out of 20 top patches correctly localized within the tumor region for both.

\begin{table}
    \caption{Evaluations of local explanations on disease localization.}
    \label{tab:disease_localization}
    \centering
    \begin{tabular}{cccccc}
        \hline
         \textbf{Model} &ABMIL &TransMIL & AdditiveMIL &CLAM &our\\
         \hline
         \textbf{Camelyon16} &0.760 & - &0.668 &0.830 & \textbf{0.871}\\
         \textbf{PANDA} &0.832 & - &\textbf{0.892} &0.852 &0.853 \\
         \hline
    \end{tabular}  
\end{table}

\subsubsection{Qualitative Evaluation}

To assess whether the concept-based local explanations align with clinical knowledge,  we conducted a user study involving three pathologists.

We selected 6 positive, 2 negative, and 2 misclassified samples from both the Camelyon16 and PANDA datasets. For each sample, we provided the pathologists with our model's WSI classification results, the top 20 patches selected by the model (see Fig.~\ref{fig:local_exp}b), and the concept list (see Table \ref{tab:conepts}). The pathologists were asked two questions for each sample. First, they were asked to select the five most significant concepts based on the provided patches. This 
question aimed to evaluate the alignment between the concepts identified by the pathologists and those predicted by the model. Second, we asked whether the pathologists agreed with the model’s classification results, given the top 20 patches, to evaluate if the explanations were useful in identifying the failure cases. This study was conducted independently with three pathologists. The agreement on concepts was quantified by the proportion of common concepts selected by both our model and the pathologists out of the top five concepts.

\begin{table}
    \caption{Average agreement between our model and the pathologists.}
    \label{tab:user_study}
    \centering
    \begin{tabular}{ccccc|ccccc}
        \hline
         \textbf{} & \multicolumn{4}{c|}{\textbf{Camelyon16}} & \textbf{} & \multicolumn{4}{c}{\textbf{PANDA}} \\
         \hline
         \textbf{} & \textbf{M} &\textbf{P1} &\textbf{P2} &\textbf{P3} & \textbf{} &\textbf{M} &\textbf{P1} &\textbf{P2} &\textbf{P3}\\
         \textbf{M} &- &0.38 &0.55 &0.53 &\textbf{M} &- &0.25 &0.45 & 0.38\\
         \textbf{P1} & - & - & 0.60& 0.60&\textbf{P1} &- &- & 0.48 & 0.48 \\
         \textbf{P2} & - & - & - & 0.65&\textbf{P2} &- &- &-& 0.50\\
         \textbf{P3} & - & - & - & - & \textbf{P3}& - & -& - & - \\
        \hline
    \multicolumn{10}{p{251pt}}{``M" denotes our model, while ``P1," ``P2," and ``P3" represent the three pathologists, respectively.}
    \end{tabular}
\end{table}

Table~\ref{tab:user_study} shows that on the Camelyon16 dataset, our model achieved agreements of 0.38, 0.55, and 0.53 with pathologists 1, 2, and 3, respectively. This means that two to three concepts among the top five predicted by our model overlapped with those selected by the pathologists. By looking into the concepts related to tumor cases, we found that ``nuclear pleomorphism'', ``disorganized cells'', ``high nuclear-to-cytoplasmic ratio'', ``loose chromatin'' and ``hyperchromatic nuclei" were the five most frequently used concepts by our model. Meanwhile, pathologists primarily referenced ``tumor cells'', ``disorganized cells'' ``enlarged nuclei'', ``high nuclear-to-cytoplasmic ratio'' and ``nuclear pleomorphism'', with 3 concepts shared between our model and the pathologists. On the PANDA dataset, the top five concepts identified by our model are ``disruption of basal cell layer'', ``glands infiltrate stroma'', ``nuclear pleomorphism'', ``apoptotic cells'' and ``neoplastic cells''. Meanwhile, the most frequently referenced concepts by pathologists included ``disruption of glandular architecture'', ``glands infiltrate stroma'', ``disruption of basal cell layer'', ``prostate glands with poorly defined borders'' and ``neoplastic cells'', again with three concepts shared between our model and the pathologists.

As shown in Table \ref{tab:user_study}, agreement scores among pathologists ranged from 0.48 to 0.65, indicating an average of two to three shared concepts between individual pathologists. This notable variance suggests that pathologists often rely on different sets of concepts for their diagnosis.

By reviewing the responses to the second question, in which we asked whether the pathologists agreed with the model's prediction, we found that for each of the four misclassified cases, at least two pathologists expressed uncertainty and suggested further examination was needed. In a false negative case from the Camelyon16 dataset, pathologists noted that the 20 patches selected by the model contained both tumor and normal tissue, therefore disagreeing with the model's ``Normal'' prediction. In a PANDA false positive case, pathologists pointed out that the selected patches were normal stroma, which did not support a ``Tumor'' prediction. The pathologists' responses to this question indicate that our model's explanations can assist users in identifying failure cases, which is desirable for human-AI collaboration.

\subsection{Evaluations of Global Explanations}

Global explanations offer insights into the model's behavior at the dataset level. After the model's training, we collected the whole slide-level concept contributions of tumor and normal predictions, calculated the mean concept vectors (Fig.~\ref{fig:global_exp}a, b), and visualized the distributions of both the concept vectors and individual concepts (Fig.~\ref{fig:global_exp}c, d, e) to assess the overall quality of the model.

\subsubsection{Quantative Evaluation}

As described in Sec.~\ref{sec:metrics}, we used the JS divergence and Silhouette score \cite{rousseeuw1987silhouettes} to evaluate the global explanations on class separability. Table \ref{tab:class_separability} presents the evaluation results for these two metrics at patch and WSI levels, with the JS Divergence score averaged across all concepts. Notably, the Silhouette scores at the WSI level are 0.518 on Camelyon16 and 0.579 on PANDA, which are significantly higher than the patch level scores of 0.216 on Camelyon16 and 0.261 on PANDA. This indicates that the concept contribution vectors at the WSI level are more tightly clustered within the same class and better separated from the opposite class. This conclusion is supported by the t-SNE \cite{van2008visualizing} plots in Fig.~\ref{fig:global_exp}c and d. In Fig.~\ref{fig:global_exp}e, we visualize the WSI-level distributions of three concepts across tumor and normal cases, with the corresponding JS divergence noted. These plots demonstrate that the individual concept has distinct distributions between tumor and normal cases.

\begin{table}[h]
    \caption{Evaluations of global explanations on class separability}
    \label{tab:class_separability}
    \centering
    \begin{tabular}{c|cc|cc}
        \hline
         \textbf{Feature} & \multicolumn{2}{c|}{\textbf{Camelyon16}} & \multicolumn{2}{c}{\textbf{PANDA}} \\
         & JS Div. & Silh. score & JS Div. & Silh. score \\
        \hline
        Top K patch concepts &0.301 &0.216 &0.291 &0.261 \\
        WSI level concepts &0.326 &0.518 &0.295 &0.579 \\
        \hline
    \end{tabular}
\end{table}

\subsubsection{Qualitative Evaluation}

In addition to the user study, we consulted one pathologist to verify whether the global explanations presented in Fig.~\ref{fig:global_exp}a, b were meaningful and could help build trust in the model. The pathologist confirmed that the top contributing concepts for tumor predictions in both datasets, such as ``high nuclear-to-cytoplasmic ratio'', ``disorganized cells'', ``nuclear pleomorphism'' for Camelyon16 and ``disruption of basal cell layer'', ``glands infiltrate stroma'', ``nuclear pleomorphism'' for PANDA, represent important pathological features for tumor detection. For normal predictions, they expected all tumor-related concepts to contribute negatively. Although not perfect, the pathologist believes our model is effective based on the concepts it uses for making predictions, and the global explanations can help to foster trust.

\section{Conclusion and Discussion}\label{sec:conclusion}

We introduced Concept MIL, an inherently interpretable model for WSI classification that offers pathology concepts as explanations. Our key idea is to use an image MIL branch to support the patch selection for an interpretable concept MIL branch, allowing predictions to be made based on a linear combination of concepts from the 20 most salient patches.

The quantitative evaluations show that inherently interpretable models like ours can be performant while offering highly interpretable explanations. Feedback from pathologists indicates that faithful local and global explanations could be helpful for building trust with the end users. Local explanations such as ours, which combine image information from WSI with high-level pathology concepts, allow pathologists to understand how a prediction is made and to investigate suspicious cases further. Global explanations that reflect the entire model's decision mechanism provide a way to assess the model’s overall quality. In the future, we plan to expand the user study to a larger group of clinicians to gain deeper insights into the clinical utility of our concept-based explanations.

Since our model uses a vision language foundation model for both image feature extraction and concept projection, a strong foundation model that effectively aligns the vision and text spaces could enhance its performance. In future work, we plan to explore the integration of pathology concepts at multiple scales, develop automatic methods for concept definition, and extend our model to support multi-class classification.

\section{Acknowledgment}
This work was funded by the Deutsche Forschungsgemeinschaft (DFG) – EXC number 2064/1 – Project number 390727645 and the Carl Zeiss Foundation in the project ``Certification and Foundations of Safe Machine Learning Systems in Healthcare". The authors thank the International Max Planck Research School for Intelligent Systems (IMPRS-IS) for supporting Susu Sun.

\bibliographystyle{plain} 
\bibliography{references}

\end{document}